\documentclass[letterpaper]{article} % DO NOT CHANGE THIS
\usepackage{aaai25}  % DO NOT CHANGE THIS
\usepackage{times}  % DO NOT CHANGE THIS
\usepackage{helvet}  % DO NOT CHANGE THIS
\usepackage{courier}  % DO NOT CHANGE THIS
\usepackage[hyphens]{url}  % DO NOT CHANGE THIS
\usepackage{graphicx} % DO NOT CHANGE THIS
\usepackage{natbib}  % DO NOT CHANGE THIS AND DO NOT ADD ANY OPTIONS TO IT
\usepackage{caption} % DO NOT CHANGE THIS AND DO NOT ADD ANY OPTIONS TO IT
\usepackage{algorithm}
\usepackage{algorithmic}

\usepackage{amsmath}
\usepackage{amssymb}
\usepackage{MnSymbol}
\usepackage{multirow}
\usepackage{booktabs}
\usepackage{ragged2e}
\usepackage{xcolor}

\usepackage{newfloat}
\usepackage{listings}
\DeclareCaptionStyle{ruled}{labelfont=normalfont,labelsep=colon,strut=off} % DO NOT CHANGE THIS
\floatstyle{ruled}
\newfloat{listing}{tb}{lst}{}
\floatname{listing}{Listing}
\title{Realistic Noise Synthesis with Diffusion Models}
\author{
    Qi Wu\textsuperscript{\rm 1}\equalcontrib,
    Mingyan Han\textsuperscript{\rm 1}\equalcontrib,
    Ting Jiang\textsuperscript{\rm 1},
    Chengzhi Jiang\textsuperscript{\rm 1},
    Jinting Luo\textsuperscript{\rm 1},
    Man Jiang\textsuperscript{\rm 1},\\
    Haoqiang Fan\textsuperscript{\rm 1},
    Shuaicheng Liu\textsuperscript{\rm 2}\thanks{Corresponding author}
}
\affiliations{
    \textsuperscript{\rm 1}Megvii Technology Inc.\\
    \textsuperscript{\rm 2}University of Electronic Science and Technology of China\\
    \{wuqiresearch, hanmy628, tjhedlen, morven126, k531756653, himandy1006\}@gmail.com,\\ fhq@megvii.com, liushuaicheng@uestc.edu.cn
}

\usepackage{bibentry}
\begin{document}

\maketitle

\begin{abstract}
Deep denoising models require extensive real-world training data, which is challenging to acquire. Current noise synthesis techniques struggle to accurately model complex noise distributions. We propose a novel Realistic Noise Synthesis Diffusor (RNSD) method using diffusion models to address these challenges. By encoding camera settings into a time-aware camera-conditioned affine modulation (TCCAM), RNSD generates more realistic noise distributions under various camera conditions. Additionally, RNSD integrates a multi-scale content-aware module (MCAM), enabling the generation of structured noise with spatial correlations across multiple frequencies. We also introduce Deep Image Prior Sampling (DIPS), a learnable sampling sequence based on depth image prior, which significantly accelerates the sampling process while maintaining the high quality of synthesized noise. Extensive experiments demonstrate that our RNSD method significantly outperforms existing techniques in synthesizing realistic noise under multiple metrics and improving image denoising performance. 
% Source code is available at \url{https://github.com/wuqi-coder/RNSD}
\end{abstract}

\begin{links}
    \link{Code}{https://github.com/wuqi-coder/RNSD}
\end{links}
\section{Introduction}
\label{sec:intro}
In deep learning, image denoising~\cite{wang2022uformer, zamir2021multi, syed2022restormer, guo2019toward, zhang2018ffdnet, kim2020transfer} is an ill-posed problem that often necessitates supervised training with extensive data pairs. A noisy image $\mathbf{y}$ in the RGB domain can be modeled as its noise-free version $\mathbf{s}$ plus noise $\mathbf{n}$ after Image Signal Processing (ISP) using the following equation:

\begin{equation}
\mathbf{y} = \textbf{ISP}(\mathbf{s} + \mathbf{n}).
\end{equation}

In contrast to the linearly modelable and spatially independent noise in RAW, noise in the RGB domain exhibits: \noindent\textbf{Irregular and Diverse Noise Distribution.} ISP post-processing parameters—such as AWB, CCM, and GAMMA—result in non-uniform noise variations across scenes, channels, ISO levels, and pixels, due to their dependence on sensor, ISO, scene, and exposure settings; \noindent\textbf{Structural and Spatial Correlation of Noise.} Spatially dependent ISP operations, including demosaicing, denoising, and sharpening, introduce local structural patterns to the noise, which increases its correlation with the signal-to-noise ratio.

\begin{figure}[t]
\centering
 \includegraphics[width=0.9\linewidth, keepaspectratio]{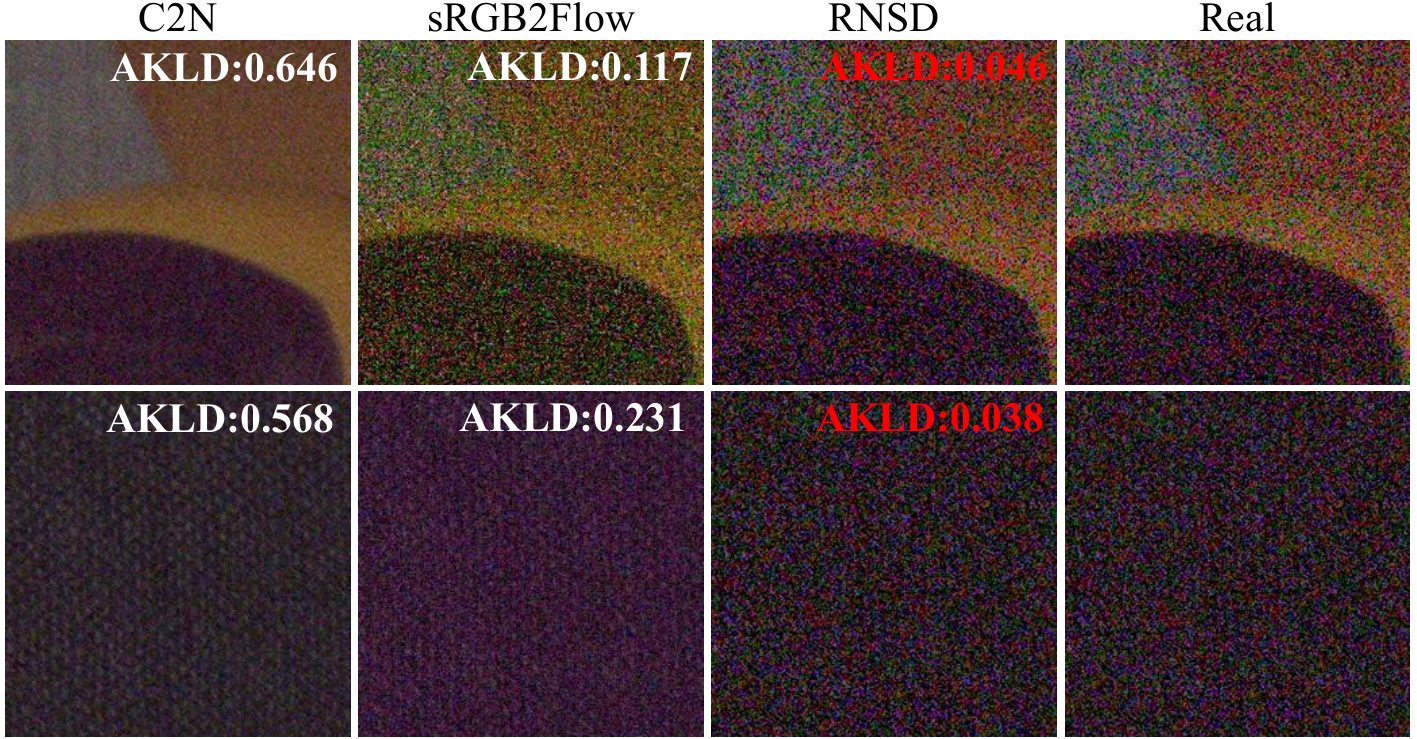}\vspace{-2mm}
\caption{Subjective results and AKLD~\cite{yue2020danet} of various noise synthesis methods, including sRGB2Flow~\cite{kousha2022modeling}, DANet~\cite{yue2020danet}, and C2N~\cite{jang2021c2n}.}
% \vspace{-4mm}
\label{fig:teaser}
\end{figure}

Most datasets~\cite{plotz2017benchmarking, xu2018real, nam2016holistic} rely on multi-frame averaging, which is not only challenging to obtain but also fails to provide diverse noise types and cannot address structural noise. Some approaches~\cite{foi2008practical, foi2009clipped, brooks2019unprocessing} model noise as Gaussian white noise, neglecting the spatial correlations present in real noise. GAN-based methods~\cite{ho2020denoising, song2020denoising,san2021noise,bansal2022cold,chen2022diffusiondet} have attempted to model real noise distributions but often struggle with instability and mode collapse due to the lack of a rigorous likelihood function, leading to a mismatch between generated and real noise distributions. In contrast, diffusion models~\cite{ho2020denoising, song2020denoising,san2021noise,bansal2022cold,chen2022diffusiondet} offer more stable and varied image generation due to their rigorous likelihood derivations. However, they have yet to be successfully applied to synthetic noise generation, possibly due to insufficient conditioning designs for complex noise distributions with spatial correlations.

In this paper, we introduce Realistic Noise Synthesize Diffusor (RNSD), a novel method for synthesizing realistic rgb noise data based on the diffusion model. RNSD has the capability to generate a large amount of noise images that closely resemble the distribution of real-world noise by clean images from various public datasets. The augmented data generated by RNSD significantly enhances the performance of existing denoising models, both in terms of noise reduction and image fidelity.
 
Specifically, RNSD uses real noisy images $\mathbf{y}$ as initial state $\mathbf{x_0}$ to build Diffusion for noise generation. To effectively accommodate diverse noise distributions, we propose a time-ware camera conditioned affine modulation, termed \textbf{TCCAM}. This module encodes varying camera settings and employs a time-adaptive conditioned affine transformation during the sampling process, permitting RNSD to synthesize diversity and realistic noise.

Additionally, we constructed a multi-scale content-aware module, \textbf{MCAM}, which integrates multi-scale guidance information of clean image into the diffusion network. This module efficiently guides the generation of signal-correlated and spatial-correlated noise. 

Based on the depth image prior that the network first learns low-frequency components and then high-frequency components, we developed Deep Image Prior Sampling (\textbf{DIPS}). Unlike DDIM, DIPS uses a distillation-based single-step model with decay sampling, reducing a 1000-step model to just 5 steps with only a 4\% accuracy loss, significantly improving sampling efficiency.

To summarize, our main contributions are as follows:

\begin{itemize}

\item We first propose a real noise data synthesis approach \textbf{RNSD} based on the diffusion model.
\item We design a time-ware camera conditioned affine modulation, \textbf{TCCAM} that can better control the distribution and level of generated noise. 

\item By constructing the multi-scale content-aware module \textbf{MCAM}, the coupling of multi-frequency information is introduced, enabling the generation of more realistic noise with spatial correlation.

\item \textbf{Deep Image Prior Sampling (DIPS)}: Leveraging the depth image prior that the network learns low-frequency before high-frequency components, DIPS improves sampling efficiency by reducing a 1000-step model to just 5 steps with only a 4\% accuracy loss, compared to DDIM.

\item Our approach achieves state-of-the-art results on multiple benchmarks and metrics, significantly enhancing the performance of denoising models.

\end{itemize}

\section{Related Work}

\noindent\textbf{Noise Models.} In digital imaging, noise sources include read noise, shot noise, fixed-pattern noise and so on. Some methods~\cite{foi2008practical, foi2009clipped} use the Gaussian-Poisson model, where read noise is signal-independent and modeled as Gaussian, while shot noise, which is signal-dependent, is modeled as Poisson noise. Shot noise is often approximated as Gaussian for simplicity~\cite{brooks2019unprocessing, guo2019toward}. However, complex ISP operations like demosaicing, tone mapping, HDR, and sharpening can disrupt noise regularity and introduce complex spatial correlations, complicating traditional noise modeling.

\noindent\textbf{GAN-based Methods.} GANs~\cite{karras2017progressive, karras2019style, brock2018large, shaham2019singan} are known for their strong data distribution fitting in image generation. Real noise can be viewed as a data distribution, leading to efforts in synthesizing noise with GANs. Jiang \textit{et al.}\cite{jang2021c2n} demonstrated that GANs can synthesize real noise using Gaussian noise inputs and unsupervised conditions. Cai \textit{et al.}\cite{cai2021learning} used a pre-training network to align generated content and noise domains. Despite their potential, GANs often face instability and poor convergence due to the absence of explicit maximum likelihood~\cite{lucic2018gans, mescheder2018training}.

\noindent\textbf{Diffusion Methods.} Diffusion models~\cite{ho2020denoising, song2020denoising, san2021noise, bansal2022cold, chen2022diffusiondet} handle the complex and diverse distribution of real noise, which can be influenced by sensor type, ISO, and ISP. Unlike GANs, diffusion models avoid mode collapse and provide more diverse results. However, they have not been effectively applied to synthetic noise generation, potentially due to the lack of conditioning designs for handling complex and varied noise distributions.

\section{Methodology}
\begin{figure*}[t]
\centering
\includegraphics[width=0.9\linewidth]{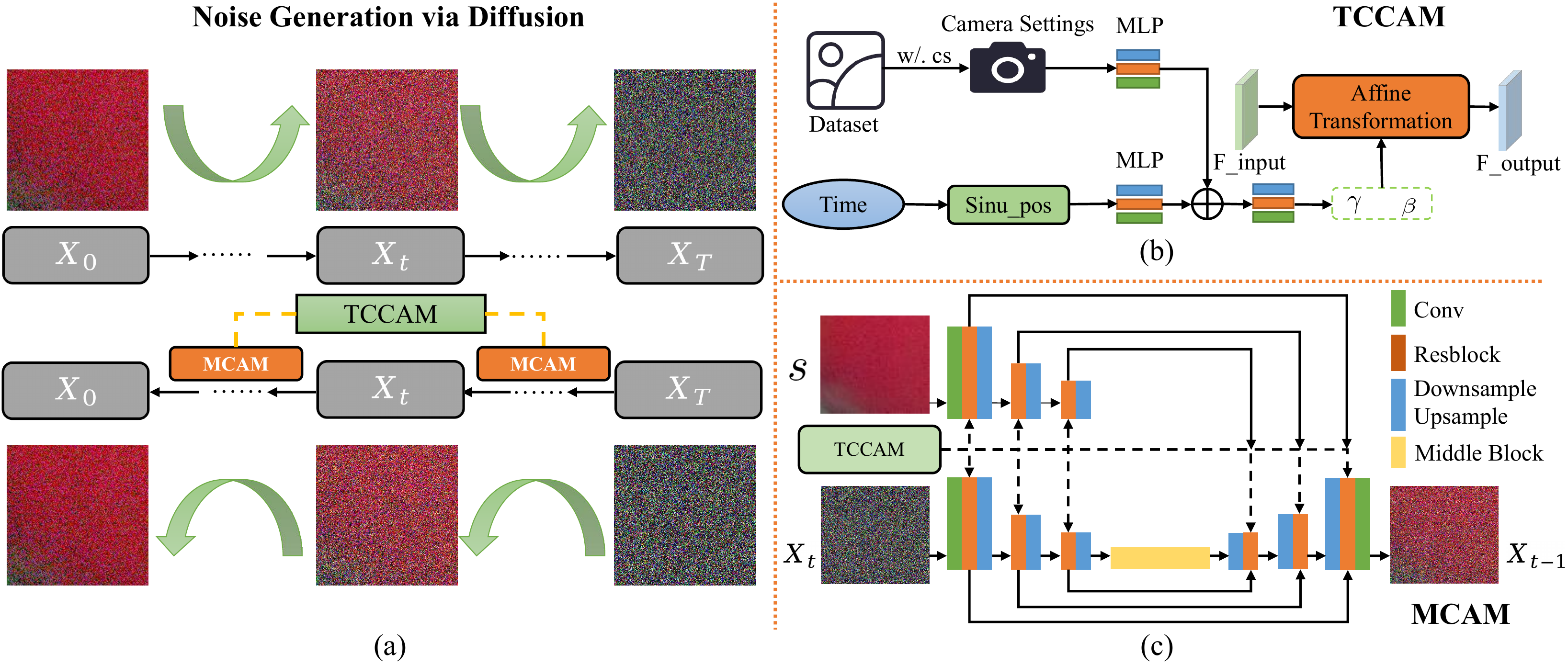}
% \vspace{-2mm}
\caption{(a) the pipeline of noise generation via diffusion. (b) The pipeline of our TCCAM. (c) The architecture of the UNet with MCAM we designed. } \label{CRNS-diffusion}
% \vspace{-5mm}
\end{figure*}

We propose a novel diffusion-based method for synthesizing realistic noisy data, termed Real Noise Synthesis via Diffusion (\textbf{RNSD}) (Fig.~\ref{CRNS-diffusion} (a)). Our method leverages real noisy images as initial conditions and incorporates a Time-aware Camera Conditioned Affine Modulation (\textbf{TCCAM}) (Fig.~\ref{CRNS-diffusion} (b)) to control the results. 
Additionally, we introduce a Multi-scale Content-aware Module (\textbf{MCAM}) (Fig.~\ref{CRNS-diffusion} (c)) to guide the generation of signal-correlated noise. Finally, we designed a learnable accelerated sampling method based on depth image prior(\textbf{DIPS})~\cite{ulyanov2018deep} as shown in algorithm~\ref{algo1}.

% \vspace{-3mm}
\subsection{Noise Generation via Diffusion}\label{gnd}
Traditional diffusion models are usually trained on noise-free style data, which can sample target domain images from any Gaussian noise distribution. In contrast, we treat images with real noise distributions as target domain images. As shown in Fig.~\ref{CRNS-diffusion} (a), by replacing $\mathbf{x}_0$ with real noise distribution data $\mathbf{y}$ and through simple settings, the diffusion model sample real noise  from any Gaussian noise distribution. 

\begin{algorithm}[t]
\caption{RNSD Training}
\label{algo2}
\begin{algorithmic}[1]
    \REPEAT
        
        \STATE $\alpha_t = 1 - \beta_t$, $\overline{\alpha}_t=\prod_{s=1}^t\alpha_s$, $t \sim \text{Uniform}(1, \dots, T)$
        \STATE $\mathbf{x_0} \sim \mathbf{q(x_0)}$, $\epsilon \sim \mathcal{N}(\mathbf{0}, \mathbf{I})$
        \STATE $\mathbf{x}_t = \sqrt{\overline{\alpha}_t}\mathbf{x}_0+\sqrt{1-\overline{\alpha}_t}\epsilon$
        % \STATE $\mathbf{\epsilon}_\theta = \textbf{MCAM}(\mathbf{x}_t, \mathbf{s},\textbf{TCCAM}(t, \mathbf{cs}))$
        \STATE Compute gradient descent step on:
        \STATE $\mathbf{\triangledown}_{\theta} || \mathbf{\epsilon} - \epsilon_\theta(\mathbf{x}_t,\textbf{MCAM}(\textbf{s}),\textbf{TCCAM}(t, \mathbf{cs})))||$
    \UNTIL{converged}
\end{algorithmic}
\end{algorithm}

\begin{algorithm}[t]
\caption{RNSD Sampling(DIPS)}
\label{algo1}
\begin{algorithmic}[1]
\STATE $\mathbf{x}_T\sim{\mathcal{N} (\mathbf{0},\mathbf{I})}$, maximum steps $T$, accelerated steps $S$\\
\STATE one-step model $\psi_\theta$, closer initial step N \\
\STATE last step $t_{last}=\frac{10}{log(S)}$, density parameter $r$
%输入条件
\FOR{$i = S,...,1$}   %FOR循环结构
% \STATE $\alpha_t=(t-1)/T$\\

\STATE $t = t_{last} + (N-t_{last})\frac{e^{r(\frac{i-1}{S-1})}-1}{e^{r} - 1}$

\STATE $t_{next} = t_{last} + (N-t_{last})\frac{e^{r(\frac{i-2}{S-1})}-1}{e^{r} - 1}$

% N - (S - i)me^{r(\frac{S-i-1}{S-2})},  r=\ln\frac{N-1}{m(S-1)}$ \\

\IF{t==N}{
\STATE $\epsilon = \psi_\theta(x_T, t, \textbf{MCAM}(\textbf{s}),\textbf{TCCAM}(t,\mathbf{cs}))$ \\
$x_{t} = \sqrt{\bar{\alpha}_{t}}(\frac{x_T - \sqrt{1 - \bar{\alpha}_{T}}\epsilon}{\sqrt{1 - \bar{\alpha}_{T}}}) + \sqrt{1 - \bar{\alpha}_{t}}\epsilon$
} 
\ENDIF

\STATE $\epsilon = \epsilon_\theta(x_t, t, \textbf{MCAM}(\textbf{s}),\textbf{TCCAM}(t, \mathbf{cs})))$ 

\IF{i==1}{
\STATE $t_{next} = 0$
}
\ENDIF

\STATE $x_{t_{next}} = \sqrt{\bar{\alpha}_{t_{next}}}(\frac{x_t - \sqrt{1 - \bar{\alpha}_{t}}\epsilon}{\sqrt{1 - \bar{\alpha}_{t}}}) + \sqrt{1 - \bar{\alpha}_{t_{next}}}\epsilon$
\ENDFOR
% \STATE 
% \STATE $\mathbf{z}\sim{\mathcal{N} (\mathbf{0},\mathbf{I})}$ if t>1, else $\mathbf{z}=\mathbf{0}$
% % \STATE $z\sim_{N(0,\mathbf{I})} if t>1, else z=0$ \\
% % \STATE $\beta, \gamma = $TCCAM$(t, cs)$ \\
% % \STATE $\mathbf{\epsilon}_\theta=$\textbf{MCAM}$(\mathbf{x}_t, \mathbf{s}, 
% %  $\textbf{TCCAM}$(t, \mathbf{cs}))$ \\

% \STATE $\mathbf{x}_{t-1} = \frac{1}{\sqrt{\alpha_t}}(\mathbf{x_{t}}-\frac{1-\alpha_t}{\sqrt{1 - {\overline{\alpha}}_t}}\epsilon_\theta(\mathbf{x}_t,\textbf{MCAM}(\textbf{s}),\textbf{TCCAM}(t, \mathbf{cs}))) + \sigma_{t}\mathbf{z} $

\STATE $\mathbf{y}=\mathbf{x}_0$
\RETURN $\mathbf{y}$
\end{algorithmic}
\end{algorithm}
% \vspace{-8mm}
% \vspace{-8mm}

Specifically, we adopt the probability model of DDPM~\cite{ho2020denoising}. In the forward process, a T-step Markov chain is used to minimize the prior probability $q(\mathbf{x}_T|\mathbf{x}_0)$, that is diffusing $\mathbf{x}_0$ to a pure Gaussian distribution $\mathbf{x}_T$ with a variance noise intensity $\beta_t$:
% \vspace{-2mm}
\begin{equation}\label{post}
\begin{aligned}
& q(\mathbf{x}_T|\mathbf{x}_0) = \prod\limits_{t=1}^Tq(\mathbf{x}_t|\mathbf{x}_{t-1}),\\
& q(\mathbf{x}_t|\mathbf{x}_{t-1}) = \mathcal{N}(\mathbf{x}_t; \sqrt{1-\beta_t}\mathbf{x}_{t-1}, \beta_t\mathbf{I}),
\end{aligned}
\end{equation}
where $\mathbf{I}$ is a unit covariance matrix. For adjacent two steps, with the help of reparameterization, $\mathbf{x}_t$ can be viewed as sampled from the prior distribution $q(\mathbf{x}_t|\mathbf{x}_{t-1})$, which is regarded as a Gaussian distribution formed by $\mathbf{x}_{t-1}$ and $\beta_t$. 

The general sampling process is obtained by inversely solving a Gaussian Markov chain process, viewed as a joint probability distribution $p_\theta(\mathbf{x}_{0:T})$, which can be understood as gradual denoising from the above Gaussian distribution $\mathbf{x}_T$ to obtain the sampled result $\mathbf{x}_0$:
% \vspace{-3mm}
\begin{equation}\label{post}
\begin{aligned}
& p_\theta(\mathbf{x}_{0:T}) = p(\mathbf{x}_T)\prod\limits_{t=1}^Tp_\theta(\mathbf{x}_{t-1}|\mathbf{x}_t), \\
& p_\theta(\mathbf{x}_{t-1}|\mathbf{x}_t) = \mathcal{N}(\mathbf{x}_{t-1};\mu_\theta(\mathbf{x}_t, t), \Sigma_t),\\
\end{aligned}
\end{equation}
where $\mu_\theta$ is the mean of the posterior distribution estimated by the network and $\Sigma_t$ is a variance obtained by $\beta_{t}$ forward calculation. We introduce additional information clean image $\mathbf{s}$ and camera settings information $\mathbf{cs}$ to make the whole process more controllable, mathematically, the process is:

\begin{equation}\label{post-condition}
\begin{aligned}
& p_\theta(\mathbf{x}_{t-1}|\mathbf{x}_t) = \mathcal{N}(\mathbf{x}_{t-1};\mathbf{\mu_t}, \Sigma_t), \\
& \mathbf{\mu_t} = \mathbf{\mu}_\theta(\mathbf{x}_t, \mathbf{s}, \mathbf{cs}, t), \\
\end{aligned}
\end{equation}

Considering the varying influence of camera settings($\mathbf{cs}$) information across different sampling steps and the spatial correlation of noise in the RGB domain, we formulate enhanced conditioning mechanisms \textbf{TCCAM}  and \textbf{MCAM} to achieve tighter coupling between the noise generation module and the diffusion-based image sampling framework.

\subsection{TCCAM: Time-aware Camera Conditioned Affine Modulation}\label{sec:dsc}

As shown in algorithm.~\ref{algo2}, normal diffusion models learn network parameters $\mathbf{\epsilon}_{\theta}$ to predict the added noise component $\epsilon_t$ from $\mathbf{x}_0$ to $\mathbf{x}_t$ during the forward process. However, real noise distributions are determined by multiple factors, including ISO gain, shutter speed, color temperature, brightness, and others. Without distinguishing noise distributions under different conditions, it is challenging to learn a generalized distribution from complex noise based on spatial photometric variations, ISO changes, and sensor variations.
The fundamental issue is that noise distributions vary significantly under different conditions. For instance, noise across different sensors can exhibit entirely different distributions. During learning, the network tends to converge to the overall expectation of the dataset, leading to fixed-mode noise patterns that cause discrepancies between generated and target noise.
To address this, we introduce five factors as shown in Algorithm~\ref{algo2}:
\begin{equation}\label{Objective function v2}
\mathbf{cs} = \mathbf{\phi}(iso, ss, st, ct, bm),
\end{equation}
where $iso$ is ISO, $ss$ is shutter speed, $st$ is sensor type, $ct$ is color temperature, and $bm$ is brightness mode. These factors are embedded using an encoding method ($\phi$) as a feature vector of camera settings $\mathbf{cs}$ to control noise generation. This explicit prior narrows the learning domain of the network, enabling it to approximate more complex and variable noise distributions.
The influence of camera settings should vary with sampling steps. For example, sensor type ($st$), strongly correlated with ISP, determines the basic form of noise and its impact is usually coupled with high-frequency information in image content. As $t$ samples from $T$ to 0, recovering image content from low to high frequency, the impact of camera settings gradually increases.
To address this, we propose a \textbf{TCCAM} with a dynamic setting mechanism, where the weights of different factors vary with sampling steps. As shown in Fig.~\ref{CRNS-diffusion} (b), the process is:

\begin{equation}\label{Objective function v3}
\begin{aligned}
& \gamma, \beta = MLP_{3}(MLP_{1}(sinu\_pos(t)) + MLP_{2}(\textbf{cs})) \\ 
& \mathbf{F}_{output} = \gamma*\mathbf{F}_{input} + \beta \\
\end{aligned}
\end{equation}
where a Multilayer Perceptron (MLP) is used to encode camera settings together with sampling steps using sinusoidal position encoding ($sinu\_{pos}$) to generate affine parameters ${\beta}$ and ${\gamma}$ at each layer of UNet. This method enables a dynamic setting influence mechanism by applying affine transformations to each layer of features $\mathbf{F}_{input}$ in UNet.

\subsection{MCAM: Multi-scale Content-aware Module}\label{sec:mfd}

Real noise distribution is inherently linked to image content, varying across brightness regions due to photon capture and ISP processing. Inspired by Zhou et al.'s\cite{zhou2020awgn} insights on noise's spatial frequency characteristics, we propose a Multi-Scale Content-aware Module (MCAM) (Fig.\ref{CRNS-diffusion}(c)) to model noise-image coupling across different frequencies.
Mathematically, our approach is as follows:

\begin{equation}
\begin{aligned}\label{Objective function v5}
& \mathbf{F}_{\mathbf{x}_{t_i}} = encoder_{i}(\mathbf{x}_t), \\
& \mathbf{F}_{\mathbf{s}_i} = encoder_{i}(\mathbf{s}), i = 1, 2, 3, \\
& \mathbf{F}_{\mathbf{o}_i} = decoder_{i}(Concat(\mathbf{F}_i, \mathbf{F}_{\mathbf{s}_i}, \mathbf{F}_{\mathbf{x}_{t_i}})),
\end{aligned}
\end{equation}
where symmetric but non-shared weight encoders are used to extract features from both $\mathbf{x}_t$ and the clean image $\mathbf{s}$ at three downsampling stages of the encoder. Alongside standard skip connections between $\mathbf{F}_i$ and $\mathbf{F}_{\mathbf{x}_{t_i}}$, we incorporate multi-scale features of $\mathbf{F}_{\mathbf{s}_i}$ in the three upsampling stages.

\begin{figure}[t]

\centering
\includegraphics[width=0.9\linewidth]{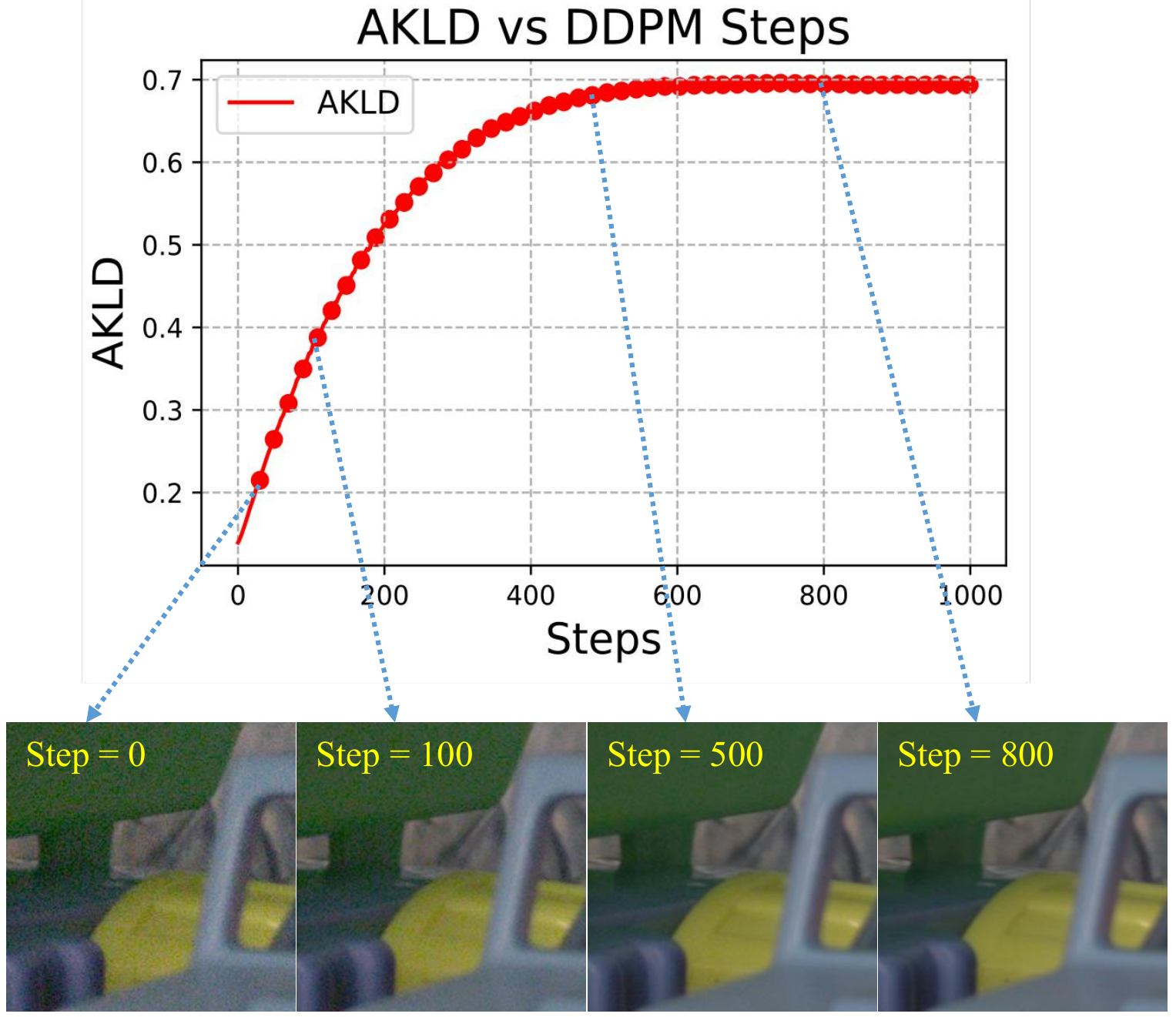}
% \vspace{-3mm}
\caption{Illustration of the motivation behind DIPS. Upper figure illustrates the variation of AKLD with respect to the steps in DDPM. Lower figure presents a selection of images generated at 0, 100, 500, and 800 steps.} \label{fig:akld_ddpm_step}
% \vspace{-7mm}
\end{figure}
\subsection{Deep Image Prior Sampling(DIPS)}\label{sec:nsg}

Based on observations from the depth image prior paper~\cite{ulyanov2018deep}, where networks first learn clean low-frequency components before high-frequency noise, we noticed a similar pattern during the DDPM sampling process. The decline of AKLD shows an increasingly large gradient as sampling progresses from 1000 steps, indicating that RNSD transitions from low-frequency content to high-frequency noise, as shown in Fig.~\ref{fig:akld_ddpm_step}. Since DDIM uses uniform sampling steps, it doesn't align well with our noise estimation task. Consequently, reducing the number of steps leads to a significant performance drop, as demonstrated in our experiments.

To address this, we propose a new sampling method:
\begin{equation}
\begin{aligned}\label{DPS_sample}
t = t_{last} + (T-t_{last})\frac{e^{r(\frac{i-1}{S-1})}-1}{e^{r} - 1}, \quad i=S:1
\end{aligned}
\end{equation}
where $T$ is the sampling step of DDPM, and $S$ is the  sampling step of DIPS. $t_{last}$ is last sampling step before step 0. Boundary effect makes last steps' generation weak, so placed in important steps as total sampling steps decrease. $r$ controls the gradient of sampling density (generally set as 
). For $T=1000$ and $S=10$, get sampling sequence [1000, 572, 327, 186, 106, 59, 33, 18, 9, 4, 0], which becomes denser as decreases.

Our basic version, DIPS (DIPS-Basic), reduces sampling steps $S$ to 30 while maintaining quality. Additionally, we find that low-frequency learning from 1000 to 200 steps can be effectively replaced by a single-step model, enabling sampling from a closer truncation step N = 200 instead of T = 1000. The specific formula for this is:
\begin{equation}
\begin{aligned}\label{DIPS_advanced}
\mathbf{\triangledown}_{\theta} || \mathbf{\psi}_\theta(x_{T}, t_{N}) - \epsilon_\theta(x_{N}, t_{N})||
\end{aligned}
\end{equation}
where a single-step model $\mathbf{\psi}_\theta$ is distilled from a pre-trained model $\mathbf{\epsilon}_\theta$, enabling us to reduce initial sampling position and build deterministic mappings to achieve 5-step sampling while maintaining quality. As follows in Algorithm \ref{algo1}, this method is referred to as \textbf{DIPS-Advanced}.

\section{Experiments}

\subsection{Experimental Settings}

\noindent\textbf{Datasets.}  To comprehensively demonstrate the effectiveness of our work, we adopt the following different datasets:

 - \textbf{SIDD}: The SIDD dataset~\cite{abdelhamed2018high}, utilized for training and evaluation, includes subsets such as SIDD small with 160 image pairs from 5 smartphone cameras, and SIDD medium with double the noise sampling. The SIDD validation set (1280 patches from unseen sensor settings) is used to test noise model and denoising models.

 % The SIDD validation set, comprising 1280 image patches from sensor settings not seen during training, is used to test denoising models.

- \textbf{DND}: DND benchmark~\cite{plotz2017benchmarking} provides 50 reference images and their realistic noisy counterparts generated using accurate sensor noise models. We use it to noise model generalization after RNSD data augmentation.

- \textbf{LSDIR}: LSDIR dataset \cite{li2023lsdir} contains 84,991 high-quality clean samples.

In all experiments, we train RNSD on SIDD small. We augment the noise sampling and scene sampling of the denoising dataset by using RNSD to generate noisy samples from the clean samples of SIDD small and 1000 randomly selected high-quality samples from LSDIR, respectively. The SIDD validation set and DND benchmark are used to assess the accuracy of RNSD noise distribution, as well as the performance and generalization of RNSD-enhanced denoising models.

%Using the SIDD validation set and DND benchmark, we evaluate the accuracy of RNSD-synthesized noise distribution and the performance and generalization of RNSD-enhanced denoising models.

%The SIDD validation set and DND benchmark are used to evaluate the accuracy of the RNSD noise distribution and the performance and generalization of RNSD-enhanced denoising models.

%, which in turn improves the performance and generalization of the denoising model.

\noindent\textbf{Evaluation and Metrics.} We use PGap and AKLD from \cite{yue2020danet} to assess noise generation. Lower PGap indicates better noise generation, while lower AKLD signifies more similar distributions. The Peak Signal-to-Noise Ratio (PSNR) and Structural Similarity Index (SSIM) are also used to evaluate the performance of the denoising model.

\begin{figure*}[t]
\centering
\includegraphics[width=0.8\linewidth]{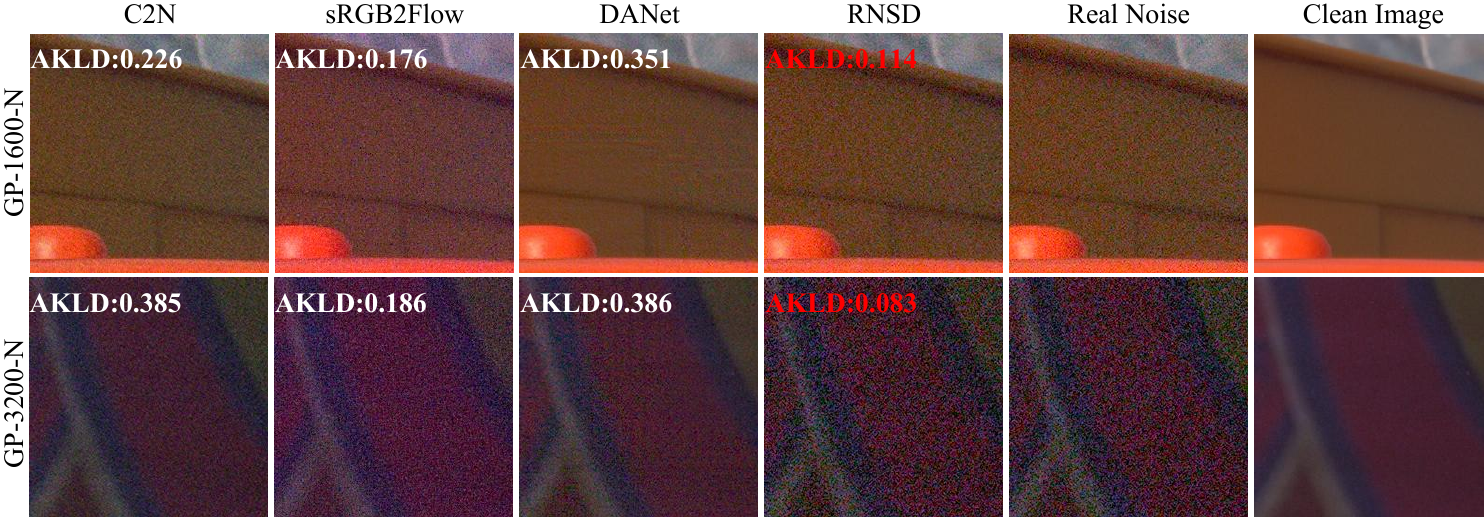}
% \vspace{-2mm}
\caption{Noisy synthesis samples from different methods, including C2N, sRGB2Flow, DANet and RNSD.  } \label{fig:compare}
% \vspace{-5mm}
\end{figure*}
%Codes on the left indicate [camera]-[ISO]-[brightness].

\noindent\textbf{Implementation Details.} We train the DDPM~\cite{ho2020denoising} diffusion model of our noise generation system with 1000 steps, a gradient accumulation step size of 2, and Adam optimizer ($\text{lr} = 8 \times 10^{-5}$). Training samples are 128$\times$128 crops from original images, with a batch size of 16. Models are trained on an NVIDIA GeForce RTX 2080 Ti GPU for $2\times10^5$ iterations.  For testing, we use Exponential Moving Average Decay (EMA) with decay 0.995. Our RNSD model achieves an inference time of 0.15 seconds to generate a batch of 16 128$\times$128 image patches using 5 sampling steps with DIPS on an NVIDIA GeForce RTX 2080 Ti GPU. Evaluation uses DIPS-Basic 30-step, matching DDPM's 1000-step accuracy. When training the denoising networks from scratch, we largely keep the original training hyper-parameters consistent with the baseline denoising models. For finetuning, we reduced the learning rate to $1 \times 10^{-6}$, maintaining other parameters, and finetuned for $1 \times 10^6$ iterations before evaluation.

\begin{table}[t]
\centering
\resizebox{1.0\columnwidth}{!}{
\setlength{\tabcolsep}{4pt}{
\begin{tabular}{c|ccccccc}
\hline
\hline
\multirow{1}{*}{Metrics}  & GRDN & C2N   & sRGB2Flow & DANet &NeCA & PNGAN & RNSD  \\ \hline
AKLD$\downarrow$                      & 0.443 & 0.314 & 0.237  & 0.212 & 0.156 & 0.153 & \textbf{0.117} \\ \hline

PGap$\downarrow$                      & 2.28 & 6.85 & 6.3     & 2.06 & 0.97 & 0.84 & \textbf{0.54} \\ \hline
\hline
\end{tabular}}}
% \vspace{-3mm}
\caption{The AKLD and PGap performances of various methods on the SIDD validation set.}
\label{tab:akld_pgap_com}
% \vspace{-3mm}
\end{table}

\begin{table}[t]
\centering
\resizebox{1.0\columnwidth}{!}{
\setlength{\tabcolsep}{4pt}{
\begin{tabular}{c|cccccc|c}
\hline
\hline
% Method                   & sRGB           \\ \hline
% Heteroscedastic Gaussian & 32.24          \\
% Isotropic Gaussian       & 32.48          \\
% Full Gaussian            & 32.72          \\
% Diagonal Gaussian        & 33.34          \\ \hline
% C2N                & 33.76          \\
% Noise Flow       & 33.81          \\
% sRGB2Flow             & 34.74          \\
% GMDCN                    & 36.03          \\
% Ours            & 36.53 \\ 
% \textbf{Ours*}          & \textbf{36.78} \\ \hline
% Real                     & 36.60           \\ \hline

% Method  & C2N  & N2F & R2F & GMDCN & Ours & Ours* & Real \\ \hline
% PSNR$\uparrow$ & 33.76 & 33.81 & 34.74 & 36.03 & 36.53 & \textbf{36.78} & 36.60 \\

\multirow{1}{*}Method  & C2N  & NoiseFlow & sRGB2Flow & GMDCN &NeCA & RNSD  & Real \\ \hline
PSNR$\uparrow$ & 33.98 & 33.81 & 34.74 & 36.07 & 37.65 & \textbf{38.11} & 38.40 \\
\hline
\hline
\end{tabular}}}
% \vspace{-4mm}
\caption{Denoising performance comparison between DnCNN trained on purely synthetic noise data and baseline. }
\label{tab:dncnn_psnr}
% \vspace{-7mm}
\end{table}

% \usepackage{multirow}
\begin{table*}[t]
\centering
\begin{center}
\resizebox{0.8\linewidth}{!}{
\setlength{\tabcolsep}{5pt} 
\begin{tabular}{c|l|cccccccc}
\cmidrule[0.5pt]{1-8}
\specialrule{-0.1em}{0pt}{0pt}
\cline{1-8}
                                    & \multicolumn{1}{c|}{}       & \multicolumn{6}{c}{SIDD validation set}                                                                                                                                                   &  &  \\ \cline{1-8}
                                    & \multicolumn{1}{c|}{Method} & \multicolumn{2}{c|}{DnCNN-B}                                   & \multicolumn{2}{c|}{RIDNet}                                    & \multicolumn{2}{c}{NAFNET}                          &  &  \\ \cline{1-8}
                                    & \multicolumn{1}{c|}{Metric} & \multicolumn{1}{c}{PSNR$\uparrow$} & \multicolumn{1}{c|}{SSIM$\uparrow$}           & \multicolumn{1}{c}{PSNR$\uparrow$} & \multicolumn{1}{c|}{SSIM$\uparrow$}           & \multicolumn{1}{c}{PSNR$\uparrow$} & \multicolumn{1}{c}{SSIM$\uparrow$} &  &  \\ \cline{1-8}
\multirow{5}{*}{\text{Increasing noise samples}} & SIDD small                  & 37.70                    & \multicolumn{1}{c|}{0.947}          & 38.20                    & \multicolumn{1}{c|}{0.951}          & 38.95                    & 0.955                   &  &  \\
                                    & SIDD small+C2N             & 38.00                         & \multicolumn{1}{c|}{0.950}               &\textcolor{blue}{38.47}                          & \multicolumn{1}{c|}{\textcolor{blue}{0.953}}               &39.01                          &0.956                          &  &  \\
                                    & SIDD small+DANet           & \textcolor{blue}{38.05}                         & \multicolumn{1}{c|}{\textcolor{blue}{0.951}}               &38.41                          & \multicolumn{1}{c|}{0.953}               &39.05                          &0.956                          &  &  \\
                                    % & SIDD small+sRGB2flow*       & 37.53                    & \multicolumn{1}{r|}{0.936}          & 38.28                    & \multicolumn{1}{r|}{0.946}          & 39.00                    & 0.954                    &  &  \\
                                    & SIDD small+sRGB2flow       & 37.79                    & \multicolumn{1}{c|}{0.950}          & 38.28                    & \multicolumn{1}{c|}{0.952}          & \textcolor{blue}{39.39}                    & \textcolor{blue}{0.957}                    &  &  \\
                                    & SIDD small+RNSD            & \textcolor{red}{38.27}           & \multicolumn{1}{c|}{\textcolor{red}{0.952}} & \textcolor{red}{38.74}           & \multicolumn{1}{c|}{\textcolor{red}{0.954}} & \textcolor{red}{39.56}           & \textcolor{red}{0.958}           &  &  \\ \cline{1-8}
\specialrule{0em}{1.0pt}{1.0pt}
\cmidrule{1-8}
\end{tabular}}
\end{center}
% \vspace{-4mm}
\caption{Denoising performance improvement brought by existing \textcolor{red}{open-source} data synthesis methods for data augmentation.}
\label{tab:improv_psnr_cvpr}
% \vspace{-6mm}
\end{table*}

\begin{figure}[t]
% \vspace{1mm}
\centering
\includegraphics[width=1.0\linewidth]{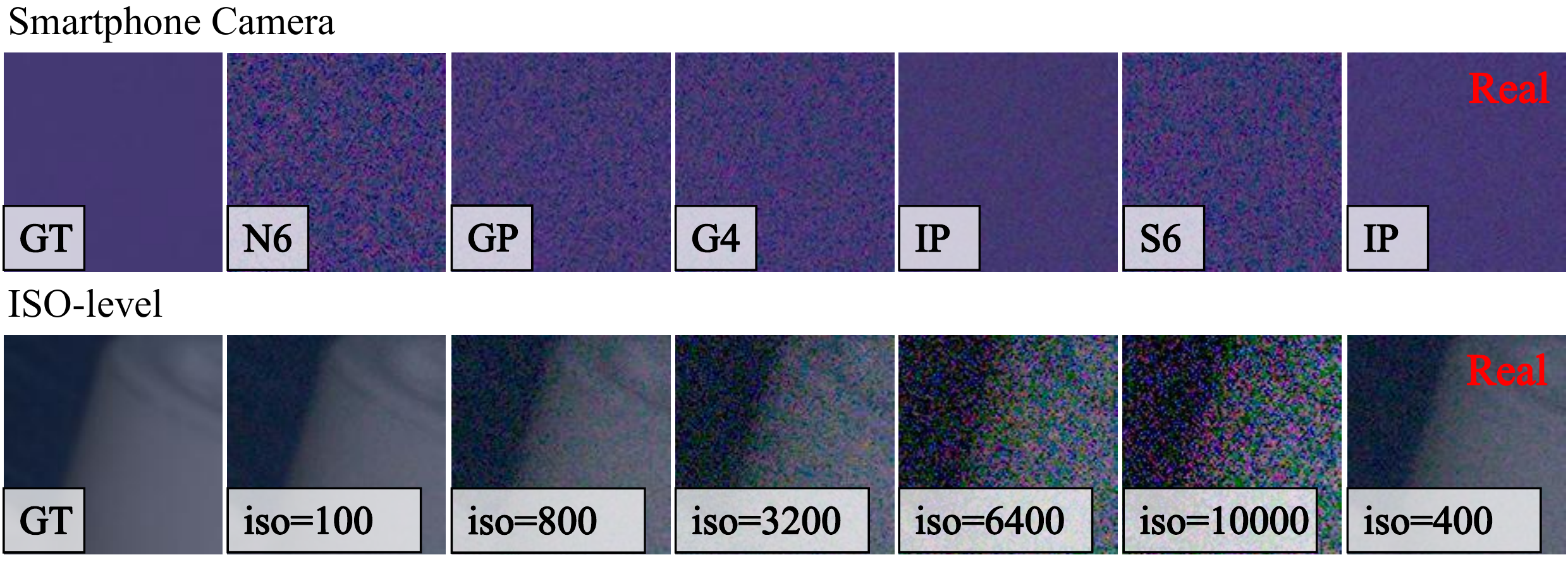}
% \vspace{-6mm}
\caption{Noise synthesis results with camera settings. } \label{fig:compare_iso}
% \vspace{-4mm}
\end{figure}
\begin{figure}[ht]
% \vspace{0mm}
\centering
\includegraphics[width=1.0\linewidth]{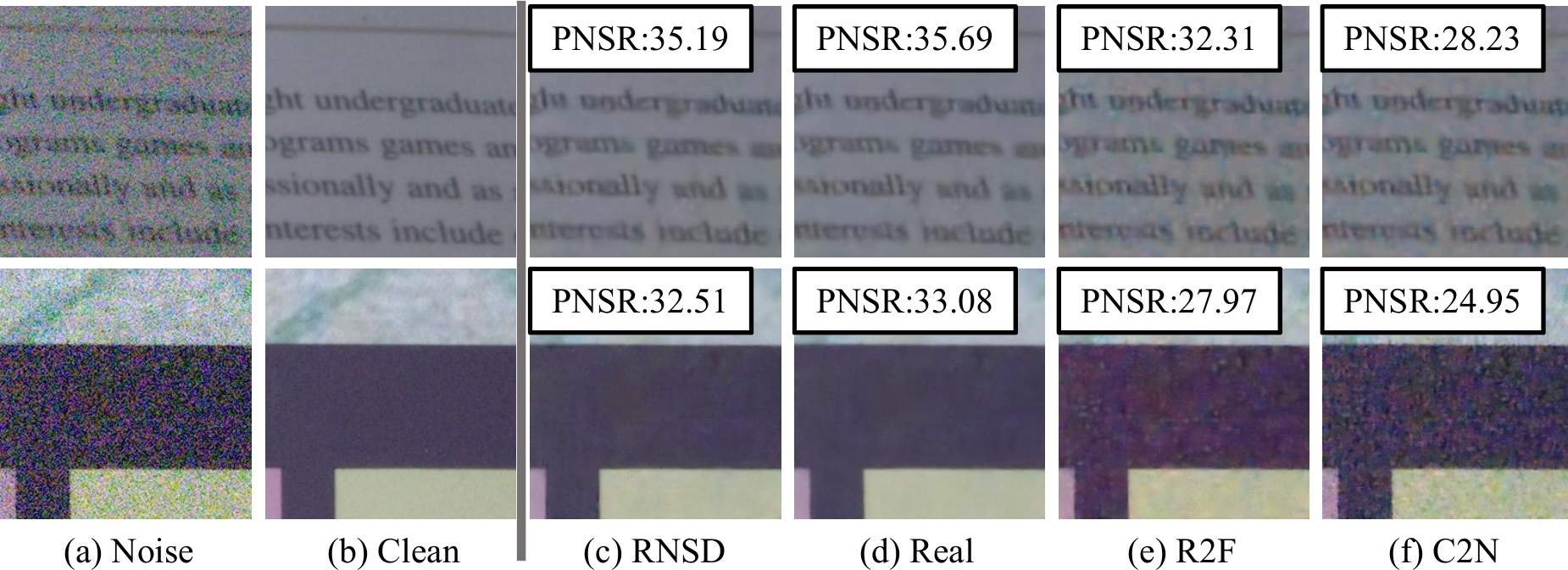}
% \vspace{-5mm}
\caption{Denoising results on SIDD validation set from DnCNN trained on noisy images from (d) real noisy of SIDD Meduim, (c) RNSD, and (e, f) two of other baselines.} 
% \caption{Denoising results on SIDD validation set from DnCNN.} 
\label{fig:compare_denoise}
% \vspace{-6mm}
\end{figure}

\begin{table}[t]
\centering
\setlength\tabcolsep{4pt} 
\resizebox{1.0\columnwidth}{!}{
\begin{tabular}{c|ccccc|cc}
\hline
\hline
     & RIDNet & MIRNet & NBNet  & Restormer & NAFNet & RIDNet* & NAFNet* \\ \hline
SIDD & 38.77  & 39.72  & 39.75    & 40.02     & 40.30  & 39.10       & \textbf{40.36}       \\ \hline
DND  & 39.29  & 39.89  & 39.89    & 40.03     & 38.43  & 39.43       & 39.05       \\ \hline
\hline
\end{tabular}}
% \vspace{-1mm}
% \caption{Comparison of PSNR results on SIDD validation set and DND benchmark for denoising methods, with * indicating fine-tuning using our RNSD noise augmentation.}
\caption{PSNR comparison on SIDD and DND, with * indicating fine-tuning using RNSD noise augmentation.}
\label{augment_clean}
% \vspace{-6mm}
\end{table}

\subsection{Qualitative Comparison}

\noindent\textbf{Visual Analysis of Noisy Images.}We compare RNSD with baselines such as C2N~\cite{jang2021c2n}, DANet~\cite{yue2020danet}, and R2F(sRGB2Flow)\cite{kousha2022modeling}, as shown in Fig.\ref{fig:compare}.  RNSD accurately mimics real-world noise patterns across sensors and ISO settings, synthesizing realistic noise while preserving color and tonal accuracy.
Fig.~\ref{fig:compare_iso} shows the controllability of RNSD in incorporating camera settings such as ISO level, shutter speed, and smartphone camera type. The noise intensity in synthesized images increases with ISO level and decreases with shutter speed. Noise patterns also vary depending on the smartphone camera type, showcasing RNSD's ability to modulate noise profiles based on camera metadata.

\noindent\textbf{Visual Analysis of Denoising Performance.} It is evident from Fig.~\ref{fig:compare_denoise} that the DnCNN~\cite{zhang2017dncnn} trained on synthetic noise samples generated by RNSD significantly outperforms other methods in terms of denoising effectiveness, closely matching the performance of the DnCNN trained with realistic noise data.

\subsection{Quantitative Comparison}

\noindent\textbf{Noise Generation.} We assess the realism of noise distributions synthesized by different methods using the public evaluation metrics AKLD and PGap on the SIDD validation set. We compare RNSD with baseline techniques including GRDN~\cite{dong2019grdn}, C2N~\cite{jang2021c2n}, sRGB2Flow~\cite{kousha2022modeling}, DANet~\cite{yue2020danet}, PNGAN~\cite{cai2021learning} and NeCA~\cite{fu2023srgb}. As shown in Table~\ref{tab:akld_pgap_com}, our method outperforms the state-of-the-art (SOTA) with a PGap reduced by 0.30 and an AKLD improved by 0.027, indicating more realistic and stable noise synthesis.

Additionally, we evaluate our method using another publicly available metric~\cite{jang2021c2n} by training the DnCNN network~\cite{zhang2017dncnn} from scratch with synthetic noise generated by RNSD. We compare its performance with C2N~\cite{jang2021c2n}, NoiseFlow~\cite{abdelhamed2019noise}, sRGB2Flow~\cite{kousha2022modeling}, GMDCN~\cite{song2023generative}, and NeCA~\cite{fu2023srgb}. As shown in Table~\ref{tab:dncnn_psnr}, our synthetic noise improves DnCNN's denoising PSNR by 0.75 dB compared to the SOTA, approaching the performance of real-data training (38.11 dB vs. 38.40 dB).

\noindent\textbf{Enhance Denoising Performance with Data Augmentation.} In the actual training process of denoising models, the performance of the model often decreases due to insufficient noise samples and limited sampling scenarios in the dataset. Our RNSD method can improve model performance by increasing noise samples and augmenting scene samples, respectively. Therefore, we verify the effectiveness of RNSD through two experimental setups.

- \textbf{Increasing noise samples}\label{Increasing_noise_samples}: SIDD Medium has twice the noise samples of SIDD Small, so we use SIDD Small to simulate insufficient sampling. RNSD is only trained on SIDD small. We augment SIDD small with different noise generation methods to train denoising models from scratch. As shown in Table.~\ref{tab:improv_psnr_cvpr}, augmenting SIDD Small with RNSD improves denoising performance, boosting PSNR/SSIM by 0.57dB/0.005, 0.54dB/0.003, and 0.61dB/0.003 for DnCNN-B, RIDNet, and NAFNet, respectively. The results validate that our synthesized diverse noise by RNSD is beneficial for training denoising models.

- \textbf{Augmenting scene samples}: We use SIDD Medium for ample noise sampling, fine-tuning RIDNet and NAFNet pre-trained on SIDD Medium with 1000 clean samples from LSDIR to augment scene samples. We evaluate these enhanced baselines on the SIDD and DND~\cite{plotz2017benchmarking} datasets. Results in Table~\ref{augment_clean} show significant performance improvements: RIDNet gains 0.33 dB and NAFNet gains 0.06 dB up to state-of-the-art levels. Notably, NAFNet shows a 0.62 dB improvement on the DND dataset. Our model, trained on a small subset of SIDD, demonstrates that augmenting training data enhances both denoising performance and generalization.

In summary, quantitative results demonstrate that RNSD effectively augments data, boosting denoising performance and generalization. This validates its ability to enhance model robustness via increased noise and sample diversity. 

\subsection{Ablation Studies}

\begin{figure}[t]
% \vspace{0mm}
\centering
\includegraphics[width=1.0\linewidth]{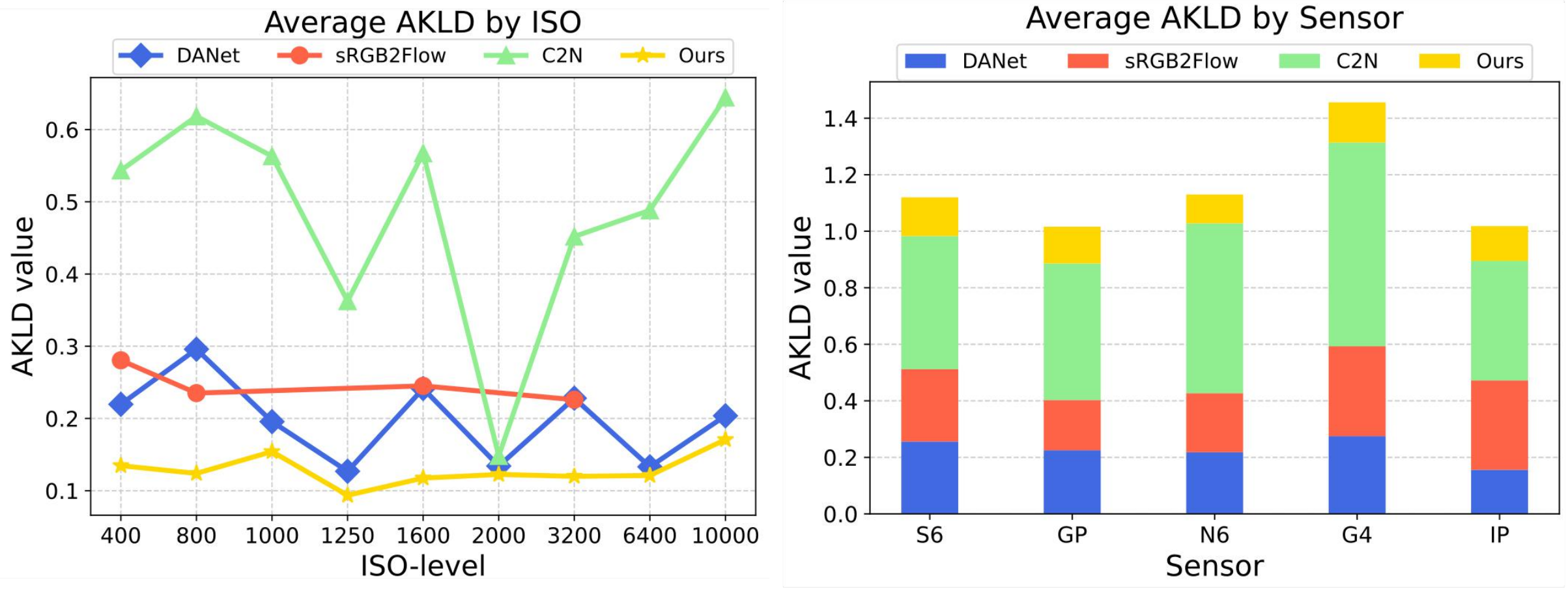}
% \includegraphics[width=1.0\linewidth]{figs/akld_iso_sensor-cropped.pdf}
% \vspace{-5mm}
% \includegraphics[width=0.5\linewidth]{figs/fig-akld_iso_sensor_curve_hist-cropped.pdf}\vspace{-2mm}
% \caption{Left: Average AKLD across ISO levels. Right: Average AKLD across sensor types. The height of each bar represents the AKLD magnitude, with shorter bars indicating better results.}
\caption{Left: Average AKLD across ISO levels. Right: Average AKLD across sensor types. }
\label{fig:curves5}
% \vspace{-7mm}
\end{figure}

\noindent\textbf{Performance Evaluation Across Different Camera Settings.} In order to evaluate the performance of RNSD across different camera settings, we consider ISO settings and sensor type as examples. Fig.~\ref{fig:curves5} shows that our noise synthesis method outperforms sRGB2Flow, C2N, and DANet in terms of lower AKLD values across all ISO levels and all sensors. This indicates our method's capability to effectively capture significant noise distribution variations and learn a more realistic noise model. 

\begin{figure}[t]
% \vspace{0mm}
\centering
\includegraphics[width=0.7\linewidth]{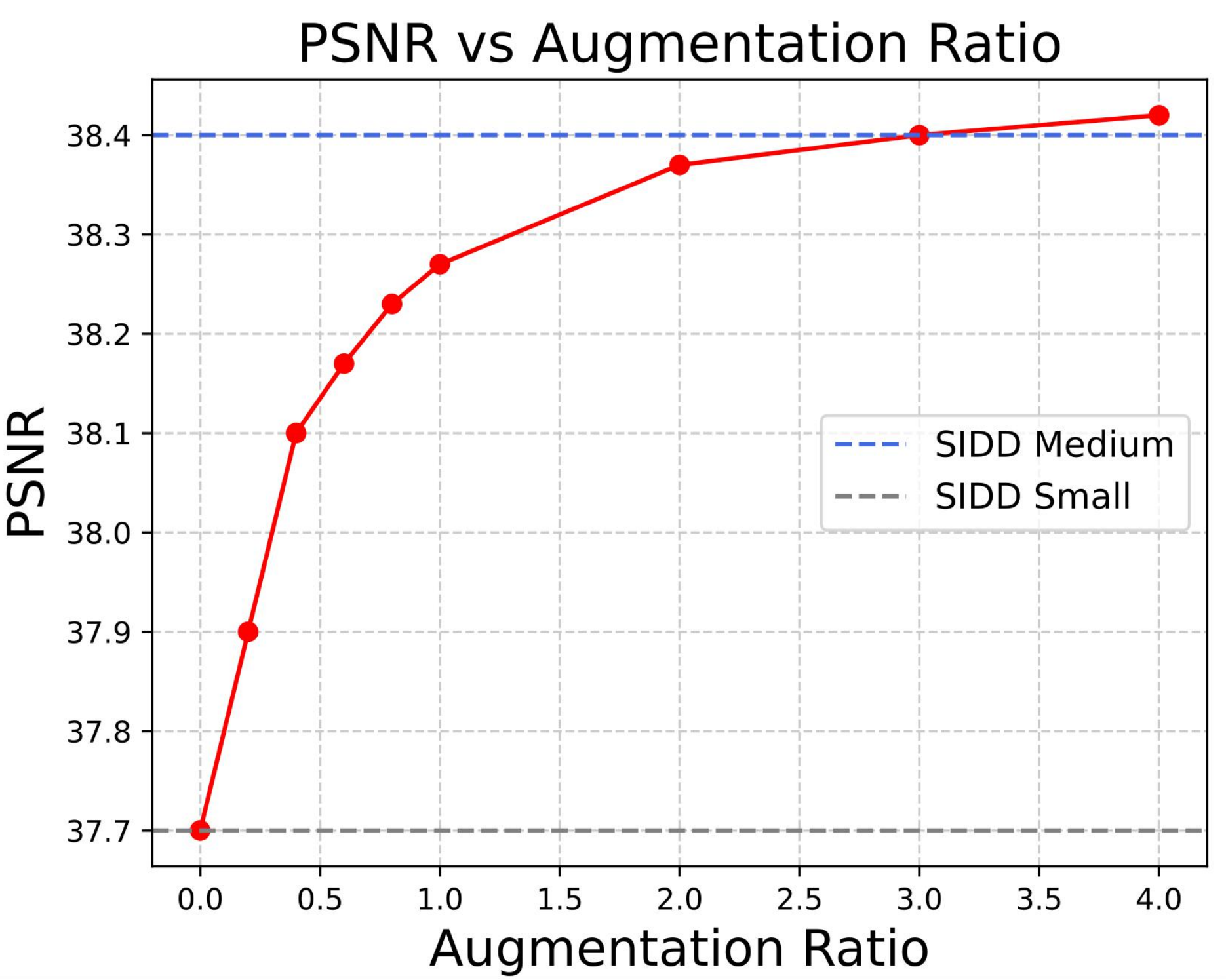}
% \vspace{-3mm}
% \includegraphics[width=0.5\linewidth]{figs/fig-akld_iso_sensor_curve_hist-cropped.pdf}\vspace{-2mm}
\caption{Noise Augmentation results at Different Ratios}
\label{fig:augmentation_psnr}
% \vspace{-5mm}
\end{figure}

\noindent\textbf{RNSD Performance on Limited Training Data.}
We assessed RNSD's performance with limited data firstly, training models on 10, 20, and 80 pairs from SIDD Small. RNSD achieved an AKLD close to DANet's (0.212) with 320 pairs when trained with fewer than 20 pairs. We used these models to synthesize additional clean-noisy pairs from SIDD, augmenting training data for DnCNN as detailed in Table ~\ref{tab:data_ratio}. Evaluation on the SIDD validation set revealed significant gains, particularly with fewer than 20 pairs, underscoring RNSD's efficacy with minimal data.

\noindent\textbf{Impact of Noise Augmentation Ratios.}
In Fig.~\ref{fig:augmentation_psnr}, we show that augmenting noise ratios from SIDD Small through RNSD progressively improves denoising accuracy. A 4x increase in training samples yields a 0.72 dB PSNR boost, highlighting RNSD's effectiveness in diversifying noise during training and enhancing model generalizability. Improvements plateau at a 3x augmentation ratio, indicating diminishing returns. Overall, RNSD's noise augmentation significantly improves real-world denoising performance.

\begin{table}[t]
\centering

% \vspace{5mm} % Adjust spacing between subtables

\begin{minipage}{1.0\columnwidth}
\centering
\resizebox{1.0\textwidth}{!}{
\setlength{\tabcolsep}{0.5mm}{
\begin{tabular}{c|ccc|c}
\hline
\hline
              & 10 pairs                           & 20 pairs        & 80 pairs   & Total (160 pairs) \\ \hline
Baseline/RNSD      & 35.49/37.08                             & 36.23/37.45             & 37.34/37.92      & 37.70/38.27  
      \\ \hline
AKLD          & 0.312                              & 0.213       & 0.178           & 0117             \\ \hline
\hline
\end{tabular}
}}
\caption{Performance vs Training Cost}
\label{tab:data_ratio}
\end{minipage}

\vspace{3mm} % Adjust spacing between subtables

\begin{minipage}{1.0\columnwidth}
\centering
\resizebox{0.6\textwidth}{!}{
\begin{tabular}{l|c}
\hline
\hline
\multicolumn{1}{c|}{Method}   & AKLD$\downarrow$           \\ \hline
\multicolumn{1}{c|}{Baseline} & 0.169          \\ \hline
+ concat camera settings      & 0.137          \\
+ TCCAM                  & 0.126           \\ 
%\hline
%+ wo MCG           & 0.134          \\
\textbf{+ MCAM}   & \textbf{0.117} \\ \hline
\hline
\end{tabular} }
\caption{Ablation study of the TCCAM and MCAM.} 
\label{tab:ablation_two_module}
\end{minipage}

\vspace{3mm}

\begin{minipage}{1.0\columnwidth}
\centering
\resizebox{1.0\textwidth}{!}{
\setlength{\tabcolsep}{3mm}{
\begin{tabular}{l|ccccc}
\hline
\hline
Steps         & 3 & 5 & 10 & 30 & 200 \\ \hline
DDIM          & 0.507   & 0.356  & 0.210   & 0.131   & 0.117   \\ \hline
DIPS-Basic     & 0.466 & 0.208  & 0.124  & \textcolor{red}{0.117}   & 0.117    \\
DIPS-Advanced & \textcolor{red}{0.138} & \textcolor{red}{0.122}  & \textcolor{red}{0.121}   & 0.120   &  0.117   \\ \hline
\hline
\end{tabular}}}
\caption{Ablation study of DIPS.}
\label{tab:dps_steps}
\end{minipage}

% \caption{Summary of ablation studies and performance analysis.}
% \label{tab:summary}
\end{table}

\noindent\textbf{RNSD's Modules.}
 % \paragraph{RNSD's Modules.}
RNSD introduces TCCAM and MCAM for realistic noise synthesis. As  shown in Table~\ref{tab:ablation_two_module}, TCCAM encodes camera settings and sampling steps into a condition tensor, reducing AKLD from 0.169 to 0.0.130, enhancing controllability. MCAM, combined with TCCAM, further reduces AKLD to 0.117, validating its role in decoupling content and camera representations. In summary, our ablation studies validate the crucial roles of RNSD's three modules in guiding realistic noise synthesis.

 \noindent\textbf{Ablation of Deep Prior Sampling (DIPS).}
We compared DIPS-Advanced, DIPS-Basic, and DDIM at various step counts (3, 5, 10, 30, 200) in Table~\ref{tab:dps_steps}. The DIPS-Advanced demonstrated remarkable efficiency, achieving results in just 5 steps with only a 4\% accuracy loss compared to the full 1000-step DDPM.  DIPS-Basic, while also effective, showed a more incremental improvement in accuracy as step counts increased, placing it second in performance. DDIM, known for its deterministic approach to sampling, was tested at reduced step counts but struggled to maintain the accuracy levels of DDPM and DPS, especially at the lower end of the step spectrum.This study shows DIPS's superior efficiency and accuracy, outperforming DDIM and making it ideal for real-time applications.

\section{Conclusion}
In this paper, we first introduce RNSD, a novel real noise synthesis method based on diffusion models. TCCAM encodes camera settings and sampling steps into time-aware affine transformation parameters, ensuring training stability and controllability while maintaining result diversity. The MCAM module uses dual multi-scale visual features to generate noise with multi-frequency spatial correlations that match real noise. Additionally, our nearly lossless accelerated sampling method, DIPS, is tailored for synthetic noise tasks. Our approach achieves state-of-the-art performance across multiple benchmarks and significantly enhances denoising models' performance and generalization in augmentation experiments.

\pdfinfo{
/TemplateVersion (2025.1)
}

\section{Acknowledgments}
This work was supported in part by National Natural Science Foundation of China (NSFC) under Grant Nos. 62372091, 62071097 and in part by Sichuan Science and Technology Program under Grant Nos. 2023NSFSC0462, 2023NSFSC0458, 2023NSFSC1972.

\bibliography{aaai25}

\begin{thebibliography}{36}
\providecommand{\natexlab}[1]{#1}

\bibitem[{Abdelhamed, Brubaker, and Brown(2019)}]{abdelhamed2019noise}
Abdelhamed, A.; Brubaker, M.~A.; and Brown, M.~S. 2019.
\newblock Noise flow: Noise modeling with conditional normalizing flows.
\newblock In \emph{{International Conference on Computer Vision}}, 3165--3173.

\bibitem[{Abdelhamed, Lin, and Brown(2018)}]{abdelhamed2018high}
Abdelhamed, A.; Lin, S.; and Brown, M.~S. 2018.
\newblock A high-quality denoising dataset for smartphone cameras.
\newblock In \emph{{IEEE Conference on Computer Vision and Pattern Recognition}}, 1692--1700.

\bibitem[{Bansal et~al.(2022)Bansal, Borgnia, Chu, Li, Kazemi, Huang, Goldblum, Geiping, and Goldstein}]{bansal2022cold}
Bansal, A.; Borgnia, E.; Chu, H.-M.; Li, J.~S.; Kazemi, H.; Huang, F.; Goldblum, M.; Geiping, J.; and Goldstein, T. 2022.
\newblock Cold diffusion: Inverting arbitrary image transforms without noise.
\newblock \emph{arXiv preprint arXiv:2208.09392}.

\bibitem[{Brock, Donahue, and Simonyan(2018)}]{brock2018large}
Brock, A.; Donahue, J.; and Simonyan, K. 2018.
\newblock Large scale GAN training for high fidelity natural image synthesis.
\newblock \emph{arXiv preprint arXiv:1809.11096}.

\bibitem[{Brooks et~al.(2019)Brooks, Mildenhall, Xue, Chen, Sharlet, and Barron}]{brooks2019unprocessing}
Brooks, T.; Mildenhall, B.; Xue, T.; Chen, J.; Sharlet, D.; and Barron, J.~T. 2019.
\newblock Unprocessing images for learned raw denoising.
\newblock In \emph{{IEEE Conference on Computer Vision and Pattern Recognition}}, 11036--11045.

\bibitem[{Cai et~al.(2021)Cai, Hu, Wang, Zhang, Pfister, and Wei}]{cai2021learning}
Cai, Y.; Hu, X.; Wang, H.; Zhang, Y.; Pfister, H.; and Wei, D. 2021.
\newblock Learning to generate realistic noisy images via pixel-level noise-aware adversarial training.
\newblock \emph{{Conference and Workshop on Neural Information Processing Systems}}, 34: 3259--3270.

\bibitem[{Chen et~al.(2022)Chen, Sun, Song, and Luo}]{chen2022diffusiondet}
Chen, S.; Sun, P.; Song, Y.; and Luo, P. 2022.
\newblock Diffusiondet: Diffusion model for object detection.
\newblock \emph{arXiv preprint arXiv:2211.09788}.

\bibitem[{Foi(2009)}]{foi2009clipped}
Foi, A. 2009.
\newblock Clipped noisy images: Heteroskedastic modeling and practical denoising.
\newblock \emph{Signal Processing}, 89(12): 2609--2629.

\bibitem[{Foi et~al.(2008)Foi, Trimeche, Katkovnik, and Egiazarian}]{foi2008practical}
Foi, A.; Trimeche, M.; Katkovnik, V.; and Egiazarian, K. 2008.
\newblock Practical Poissonian-Gaussian noise modeling and fitting for single-image raw-data.
\newblock \emph{{IEEE Transactions on Image Processing}}, 17(10): 1737--1754.

\bibitem[{Fu, Guo, and Wen(2023)}]{fu2023srgb}
Fu, Z.; Guo, L.; and Wen, B. 2023.
\newblock sRGB Real Noise Synthesizing with Neighboring Correlation-Aware Noise Model.
\newblock In \emph{{IEEE Conference on Computer Vision and Pattern Recognition}}, 1683--1691.

\bibitem[{Guo et~al.(2019)Guo, Yan, Zhang, Zuo, and Zhang}]{guo2019toward}
Guo, S.; Yan, Z.; Zhang, K.; Zuo, W.; and Zhang, L. 2019.
\newblock Toward convolutional blind denoising of real photographs.
\newblock In \emph{{IEEE Conference on Computer Vision and Pattern Recognition}}, 1712--1722.

\bibitem[{Ho, Jain, and Abbeel(2020)}]{ho2020denoising}
Ho, J.; Jain, A.; and Abbeel, P. 2020.
\newblock Denoising diffusion probabilistic models.
\newblock \emph{{Conference and Workshop on Neural Information Processing Systems}}, 33: 6840--6851.

\bibitem[{Jang et~al.(2021)Jang, Lee, Son, and Lee}]{jang2021c2n}
Jang, G.; Lee, W.; Son, S.; and Lee, K.~M. 2021.
\newblock C2n: Practical generative noise modeling for real-world denoising.
\newblock In \emph{{IEEE Conference on Computer Vision and Pattern Recognition}}, 2350--2359.

\bibitem[{Karras et~al.(2017)Karras, Aila, Laine, and Lehtinen}]{karras2017progressive}
Karras, T.; Aila, T.; Laine, S.; and Lehtinen, J. 2017.
\newblock Progressive growing of gans for improved quality, stability, and variation.
\newblock \emph{arXiv preprint arXiv:1710.10196}.

\bibitem[{Karras, Laine, and Aila(2019)}]{karras2019style}
Karras, T.; Laine, S.; and Aila, T. 2019.
\newblock A style-based generator architecture for generative adversarial networks.
\newblock In \emph{{IEEE Conference on Computer Vision and Pattern Recognition}}, 4401--4410.

\bibitem[{Kim, Chung, and Jung(2019)}]{dong2019grdn}
Kim, D.-W.; Chung, J.~R.; and Jung, S.-W. 2019.
\newblock GRDN:Grouped Residual Dense Network for Real Image Denoising and GAN-based Real-world Noise Modeling.
\newblock In \emph{{IEEE Conference on Computer Vision and Pattern Recognition Workshop}}, 0--0.

\bibitem[{Kim et~al.(2020)Kim, Soh, Park, and Cho}]{kim2020transfer}
Kim, Y.; Soh, J.~W.; Park, G.~Y.; and Cho, N.~I. 2020.
\newblock Transfer learning from synthetic to real-noise denoising with adaptive instance normalization.
\newblock In \emph{{IEEE Conference on Computer Vision and Pattern Recognition}}, 3482--3492.

\bibitem[{Kousha et~al.(2022)Kousha, Maleky, Brown, and Brubaker}]{kousha2022modeling}
Kousha, S.; Maleky, A.; Brown, M.~S.; and Brubaker, M.~A. 2022.
\newblock Modeling sRGB Camera Noise with Normalizing Flows.
\newblock In \emph{{IEEE Conference on Computer Vision and Pattern Recognition}}, 17463--17471.

\bibitem[{Li et~al.(2023)Li, Zhang, Liang, Cao, Liu, Gong, Zhang, Tang, Liu, Demandolx, Ranjan, Timofte, and Van~Gool}]{li2023lsdir}
Li, Y.; Zhang, K.; Liang, J.; Cao, J.; Liu, C.; Gong, R.; Zhang, Y.; Tang, H.; Liu, Y.; Demandolx, D.; Ranjan, R.; Timofte, R.; and Van~Gool, L. 2023.
\newblock LSDIR Dataset: A Large Scale Dataset for Image Restoration.

\bibitem[{Lucic et~al.(2018)Lucic, Kurach, Michalski, Gelly, and Bousquet}]{lucic2018gans}
Lucic, M.; Kurach, K.; Michalski, M.; Gelly, S.; and Bousquet, O. 2018.
\newblock Are gans created equal? a large-scale study.
\newblock \emph{{Conference and Workshop on Neural Information Processing Systems}}, 31.

\bibitem[{Mescheder, Geiger, and Nowozin(2018)}]{mescheder2018training}
Mescheder, L.; Geiger, A.; and Nowozin, S. 2018.
\newblock Which training methods for GANs do actually converge?
\newblock In \emph{International conference on machine learning}, 3481--3490. PMLR.

\bibitem[{Nam et~al.(2016)Nam, Hwang, Matsushita, and Kim}]{nam2016holistic}
Nam, S.; Hwang, Y.; Matsushita, Y.; and Kim, S.~J. 2016.
\newblock A holistic approach to cross-channel image noise modeling and its application to image denoising.
\newblock In \emph{{IEEE Conference on Computer Vision and Pattern Recognition}}, 1683--1691.

\bibitem[{Plotz and Roth(2017)}]{plotz2017benchmarking}
Plotz, T.; and Roth, S. 2017.
\newblock Benchmarking denoising algorithms with real photographs.
\newblock In \emph{{IEEE Conference on Computer Vision and Pattern Recognition}}, 1586--1595.

\bibitem[{San-Roman, Nachmani, and Wolf(2021)}]{san2021noise}
San-Roman, R.; Nachmani, E.; and Wolf, L. 2021.
\newblock Noise estimation for generative diffusion models.
\newblock \emph{arXiv preprint arXiv:2104.02600}.

\bibitem[{Shaham, Dekel, and Michaeli(2019)}]{shaham2019singan}
Shaham, T.~R.; Dekel, T.; and Michaeli, T. 2019.
\newblock Singan: Learning a generative model from a single natural image.
\newblock In \emph{{IEEE Conference on Computer Vision and Pattern Recognition}}, 4570--4580.

\bibitem[{Song, Meng, and Ermon(2020)}]{song2020denoising}
Song, J.; Meng, C.; and Ermon, S. 2020.
\newblock Denoising diffusion implicit models.
\newblock \emph{arXiv preprint arXiv:2010.02502}.

\bibitem[{Song et~al.(2023)Song, Zhang, Ayd{\i}n, Mansour, and Schroers}]{song2023generative}
Song, M.; Zhang, Y.; Ayd{\i}n, T.~O.; Mansour, E.~A.; and Schroers, C. 2023.
\newblock A Generative Model for Digital Camera Noise Synthesis.
\newblock \emph{arXiv preprint arXiv:2303.09199}.

\bibitem[{Ulyanov, Vedaldi, and Lempitsky(2018)}]{ulyanov2018deep}
Ulyanov, D.; Vedaldi, A.; and Lempitsky, V. 2018.
\newblock Deep image prior.
\newblock In \emph{{IEEE Conference on Computer Vision and Pattern Recognition}}, 9446--9454.

\bibitem[{Wang et~al.(2022)Wang, Cun, Bao, Zhou, Liu, and Li}]{wang2022uformer}
Wang, Z.; Cun, X.; Bao, J.; Zhou, W.; Liu, J.; and Li, H. 2022.
\newblock Uformer: A General U-Shaped Transformer for Image Restoration.
\newblock In \emph{{IEEE Conference on Computer Vision and Pattern Recognition}}, 17683--17693.

\bibitem[{Xu et~al.(2018)Xu, Li, Liang, Zhang, and Zhang}]{xu2018real}
Xu, J.; Li, H.; Liang, Z.; Zhang, D.; and Zhang, L. 2018.
\newblock Real-world noisy image denoising: A new benchmark.
\newblock \emph{arXiv preprint arXiv:1804.02603}.

\bibitem[{Yue et~al.(2020)Yue, Zhao, Zhang, and Meng}]{yue2020danet}
Yue, Z.; Zhao, Q.; Zhang, L.; and Meng, D. 2020.
\newblock Dual Adversarial Network: Toward Real-world Noise Removal and Noise Generation.
\newblock In \emph{{European Conference on Computer Vision}}, 41--58.

\bibitem[{Zamir et~al.(2022)Zamir, Arora, Khan, Hayat, Khan, and Yang}]{syed2022restormer}
Zamir, S.~W.; Arora, A.; Khan, S.; Hayat, M.; Khan, F.~S.; and Yang, M.-H. 2022.
\newblock Restormer: Efficient Transformer for High-Resolution Image Restoration.
\newblock In \emph{{IEEE Conference on Computer Vision and Pattern Recognition}}, 5728--5739.

\bibitem[{Zamir et~al.(2021)Zamir, Arora, Khan, Hayat, Khan, Yang, and Shao}]{zamir2021multi}
Zamir, S.~W.; Arora, A.; Khan, S.; Hayat, M.; Khan, F.~S.; Yang, M.-H.; and Shao, L. 2021.
\newblock Multi-stage progressive image restoration.
\newblock In \emph{{IEEE Conference on Computer Vision and Pattern Recognition}}, 14821--14831.

\bibitem[{Zhang et~al.(2017)Zhang, Zuo, Chen, Meng, and Zhang}]{zhang2017dncnn}
Zhang, K.; Zuo, W.; Chen, Y.; Meng, D.; and Zhang, L. 2017.
\newblock Beyond a Gaussian Denoiser: Residual Learning of Deep CNN for Image Denoising.
\newblock In \emph{{IEEE Transactions on Image Processing}}, 3142--3155.

\bibitem[{Zhang, Zuo, and Zhang(2018)}]{zhang2018ffdnet}
Zhang, K.; Zuo, W.; and Zhang, L. 2018.
\newblock FFDNet: Toward a fast and flexible solution for CNN-based image denoising.
\newblock \emph{{IEEE Transactions on Image Processing}}, 27(9): 4608--4622.

\bibitem[{Zhou et~al.(2020)Zhou, Jiao, Huang, Wang, Wang, Shi, and Huang}]{zhou2020awgn}
Zhou, Y.; Jiao, J.; Huang, H.; Wang, Y.; Wang, J.; Shi, H.; and Huang, T. 2020.
\newblock When awgn-based denoiser meets real noises.
\newblock In \emph{{AAAI Conference on Artificial Intelligence}}, volume~34, 13074--13081.

\end{thebibliography}

\end{document}